\documentclass[10pt,twocolumn,letterpaper]{article}

\usepackage{iccv}
\usepackage{times}
\usepackage{epsfig}
\usepackage{graphicx}
\usepackage{amsmath}
\usepackage{amssymb}
\usepackage{url}
\usepackage{multirow}
\usepackage{adjustbox}
\usepackage{threeparttable}

\usepackage[breaklinks=true,bookmarks=false]{hyperref}

\iccvfinalcopy % *** Uncomment this line for the final submission

\def\iccvPaperID{3832} % *** Enter the ICCV Paper ID here

\ificcvfinal\fi

\begin{document}

%%%%%%%%% TITLE
\title{STM: SpatioTemporal and Motion Encoding for Action Recognition}

\author{Boyuan Jiang \thanks{The work was done during an internship at SenseTime.}\\
Zhejiang University\\
{\tt\small byjiang@zju.edu.cn}
\and
MengMeng Wang \thanks{Corresponding author.}\\
SenseTime Group Limited\\
{\tt\small wangmengmeng@sensetime.com}
\and
Weihao Gan\\
SenseTime Group Limited\\
{\tt\small ganweihao@sensetime.com}
\and
Wei Wu\\
SenseTime Group Limited\\
{\tt\small wuwei@sensetime.com}
\and
Junjie Yan\\
SenseTime Group Limited\\
{\tt\small yanjunjie@sensetime.com}
}

\maketitle
\ificcvfinal\thispagestyle{empty}\fi

%%%%%%%%% ABSTRACT
\begin{abstract}
Spatiotemporal and motion features are two complementary and crucial information for video action recognition. Recent state-of-the-art methods adopt a 3D CNN stream to learn spatiotemporal features and another flow stream to learn motion features. In this work, we aim to efficiently encode these two features in a unified 2D framework. To this end, we first propose an STM block, which contains a Channel-wise SpatioTemporal Module (CSTM) to present the spatiotemporal features and a Channel-wise Motion Module (CMM) to efficiently encode motion features. We then replace original residual blocks in the ResNet architecture with STM blcoks to form a simple yet effective STM network by introducing very limited extra computation cost. Extensive experiments demonstrate that the proposed STM network outperforms the state-of-the-art methods on both temporal-related datasets (i.e., Something-Something v1 \& v2 and Jester) and scene-related datasets (i.e., Kinetics-400, UCF-101, and HMDB-51) with the help of encoding spatiotemporal and motion features together.
\end{abstract}

%%%%%%%%% BODY TEXT
\section{Introduction}
\begin{figure}[t]
	\begin{center}
		\includegraphics[width=\columnwidth]{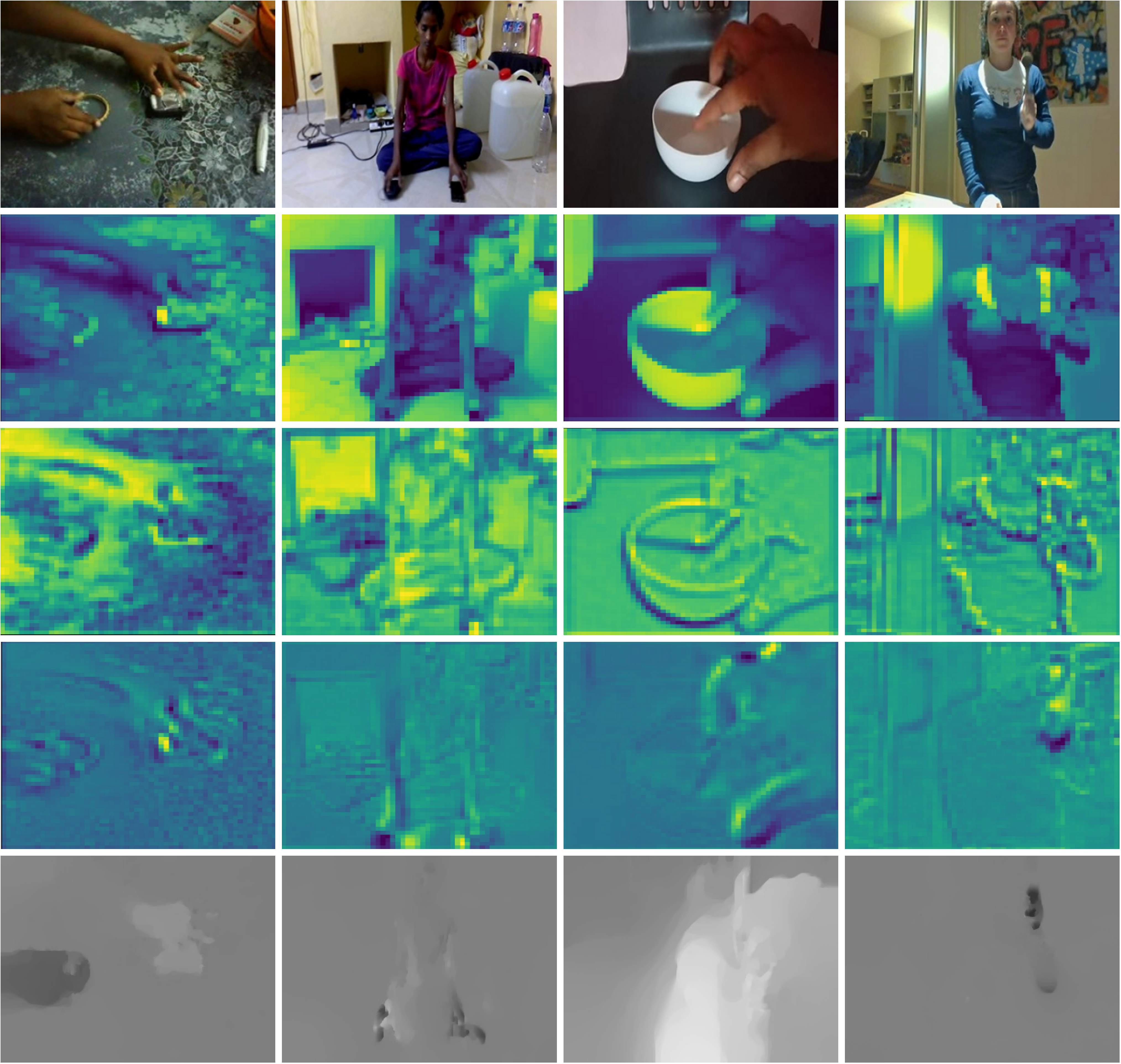}
	\end{center}
	\caption{Feature visualization of STM block. First row is the input frames. Second row is the input feature maps of Conv2\_1 block. Third row is the output spatiotemporal feature maps of CSTM. The fourth row is the output motion feature maps of CMM. The last row is the optical flow extracted by TV-L1.}
	\label{feature map}
\end{figure}
Following the rapid development of the cloud and edge computing, we are used to engaged in social platforms and live under the cameras. In the meanwhile, various industries, such as in security and transportation, collect vast amount of videos which contain a wealth of information, ranging from people's behavior, traffic, and etc. Huge video information attracts more and more researchers to the video understanding field. The first step of the video understanding is action recognition which aims to recognize the human actions in videos. The most important features for action recognition are the spatiotemporal and motion features where the former encodes the relationship of spatial features from different timestamps while the latter presents motion features between neighboring frames.
	
The existing methods for action recognition can be summarized into two categories. The first type is based on two-stream neural networks \cite{feichtenhofer2016convolutional, wang2016temporal, wang2017spatiotemporal, feichtenhofer2017spatiotemporal}, which consists of an RGB stream with RGB frames as input and a flow stream with optical flow as input. The spatial stream models the appearance features (not spatiotemporal features) without considering the temporal information. The flow stream is usually called as a temporal stream, which is designed to model the temporal cues. However, we argue that it is inaccurate to refer the flow stream as the temporal stream because the optical flow only represent the motion features between the neighboring frames and the structure of this stream is almost the same to the spatial stream with 2D CNN. Therefore, this flow stream lacks of the ability to capture the long-range temporal relationship. Besides, the extraction of optical flow is expensive in both time and space, which limits vast industrial applications in the real world.
	
	The other category is the 3D convolutional networks (3D CNNs) based methods, which is designed to capture the spatiotemporal features\cite{tran2015learning, carreira2017quo, stroud2018d3d, chen20182}. 3D convolution is able to represent the temporal features as well as the spatial features together benefiting from the extended temporal dimension. With stacked 3D convolutions, 3D CNNs can capture long-range temporal relationship. Recently, the optimization of this framework with tremendous parameters becomes popular because of the release of large-scale video datasets such as Kinetics \cite{carreira2017quo}. With the help of pre-training on large-scale video datasets, 3D CNN based methods have achieved superior performance to 2D CNN based methods. However, although 3D CNN can model spatiotemporal information from RGB inputs directly, many methods \cite{varol2018long,carreira2017quo} still integrate an independent optical-flow motion stream to further improve the performance with motion features. Therefore, these two features are complementary to each other in action recognition. Nevertheless, expanding the convolution kernel from 2D to 3D and the two-stream structure will inevitably increase the computing cost by an order of magnitude, which limits its real applications.
	
	Inspired by the above observation, we propose a simple yet effective method referred as STM network, to integrate both SpatioTemporal and Motion features in a unified 2D CNN framework, without any 3D convolution and optical flow pre-calculation. Given an input feature map, we adopt a Channel-wise Spatiotemporal Module (CSTM) to present the spatiotemporal features and a Channel-wise Motion Module (CMM) to encode the motion features. We also insert an identity mapping path to combine them together as a block named STM block. The STM blocks can be easily inserted into existing ResNet \cite{he2016deep} architectures by replacing the original residual blocks to form the STM networks with negligible extra parameters. As shown in Fig. \ref{feature map}, we visualize our STM block with CSTM and CMM features. The CSTM has learned the spatiotemporal features which pay more attention on the main object parts of the action interaction compared to the original input features. As for the CMM, it captures the motion features with the distinct edges just like optical flow. The main contributions of our work can be summarized as follows:
	\begin{itemize}
		\item We propose a Channel-wise Spatiotemporal Module (CSTM) and a Channel-wise Motion Module (CMM) to encode the complementary spatiotemporal and motion features in a unified 2D CNN framework.
		\item A simple yet effective network referred as STM Network is proposed with our STM blocks, which can be inserted into existing ResNet architecture by introducing very limited extra computation cost.
		\item Extensive experiments demonstrate that by integrating both spatiotemporal and motion features together, our method outperforms the state-of-the-art methods on several public benchmark datasets including Something-Something\cite{goyal2017something}, Kinetics \cite{carreira2017quo}, Jester \cite{jester}, UCF101 \cite{soomro2012ucf101} and HMDB-51 \cite{Kuehne11}.
	\end{itemize}

	%-------------------------------------------------------------------------
	\section{Related Works}
	% 	\noindent \textbf{2D CNN based methods.}
	With the great success of deep convolution networks in the computer vision area, a large number of CNN-based methods have been proposed for action recognition and have gradually surpassed the performance of traditional methods \cite{wang2011action,wang2013action}. A sequence of advances adopt 2D CNNs as the backbone and classify a video by simply aggregating frame-wise prediction \cite{KarpathyCVPR14}. However, these methods only model the appearance feature of each frame independently while ignore the dynamics between frames, which results in inferior performance when recognizing temporal-related videos. To handle the mentioned drawback, two-stream based methods \cite{feichtenhofer2016convolutional, wang2016temporal, wang2017spatiotemporal, chen20182, feichtenhofer2017spatiotemporal} are introduced by modeling appearance and dynamics separately with two networks and fuse two streams through middle or at last. Among these methods, Simonyan et al. \cite{simonyan2014two} first proposed the two-stream ConvNet architecture with both spatial and temporal networks. Temporal Segment Networks (TSN) \cite{wang2016temporal} proposed a sparse temporal sampling strategy for the two-stream structure and fused the two streams by a weighted average at the end. Feichtenhofer et al. \cite{feichtenhofer2016spatiotemporal,feichtenhofer2017spatiotemporal} studied the fusion strategies in the middle of the two streams in order to obtain the spatiotemporal features. However, these types of methods mainly suffer from two limitations. First, these methods need pre-compute optical flow, which is expensive in both time and space. Second, the learned feature and final prediction from multiple segments are fused simply using weighted or average sum, making it inferior to temporal-relationship modeling.
	% 	\noindent \textbf{3D CNN based methods.}
	
	Another type of methods tries to learn spatiotemporal features from RGB frames directly with 3D CNN \cite{tran2015learning,carreira2017quo,diba2018spatio,feichtenhofer2018slowfast, stroud2018d3d}. C3D \cite{tran2015learning} is the first work to learn spatiotemporal features using deep 3D CNN. However, with tremendous parameters to be optimized and lack of high-quality large-scale datasets, the performance of C3D remains unsatisfactory. I3D \cite{carreira2017quo} inflated the ImageNet pre-trained 2D kernel into 3D to capture spatiotemporal features and modeled motion features with another flow stream. I3D has achieved very competitive performance in benchmark datasets with the help of high-quality large-scale Kinetics dataset and the two-stream setting. Since 3D CNNs try to learn local correlation along the input channels, STCNet \cite{diba2018spatio} inserted its STC block into 3D ResNet to captures both spatial-channels and temporal-channels correlation information throughout network layers. Slowfast \cite{feichtenhofer2018slowfast} involved a slow path to capture spatial semantics and a fast path to capture motion at fine temporal resolution. Although 3D CNN based methods have achieved state-of-the-art performance, they still suffer from heavy computation, making it hard to deploy in real-world applications.
	
	To handle the heavy computation of 3D CNNs, several methods are proposed to find the trade-off between precision and speed \cite{tran2018closer,xie2018rethinking,zolfaghari2018eco,zhou2018mict,sun2015human,qiu2017learning}. Tran et al. \cite{tran2018closer} and Xie et al. \cite{xie2018rethinking} discussed several forms of spatiotemporal convolutions including employing 3D convolution in early layers and 2D convolution in deeper layers (bottom-heavy) or reversed the combinations (top-heavy). P3D \cite{qiu2017learning} and R(2+1)D \cite{tran2018closer} tried to reduce the cost of 3D convolution by decomposing it into 2D spatial convolution and 1D temporal convolution.  TSM \cite{lin2018temporal} further introduced the temporal convolution by shifting part of the channels along the temporal dimension. Our proposed CSTM branch is similar to these methods in the mean of learning spatiotemporal features, while we employ channel-wise 1D convolution to capture different temporal relationship for different channels. Though these methods are successful in balancing the heavy computation of 3D CNNs, they inevitably need the help of two-stream networks with a flow stream to incorporate the motion features to obtain their best performance. Motion information is the key difference between video-based recognition and image-based recognition task. However, calculating optical flow with TV-L1 method \cite{zach2007duality} is expensive in both time and space. Recently many approaches have been proposed to estimate optical flow with CNN \cite{dosovitskiy2015flownet,ilg2017flownet,fan2018end,ranjan2017optical} or explored alternatives of optical flow \cite{wang2016temporal, zhao2018recognize, sun2018optical, lee2018motion}. TSN frameworks \cite{wang2016temporal} involved RGB difference between two frames to represent motion in videos. Zhao et al. \cite{zhao2018recognize} used cost volume processing to model apparent motion. Optical Flow guided Feature (OFF) \cite{sun2018optical} contains a set of operators including sobel and element-wise subtraction for OFF generation. MFNet \cite{lee2018motion} adopted five fixed motion filters as a motion block to find feature-level temporal features between two adjacent time steps. Our proposed CMM branch is also designed for finding better yet lightweight alternative motion representation. The main difference is that we learn different motion features for different channels for every two adjacent time steps.
	
	% 	\noindent \textbf{Channel-wise convolution.} Channel-wise convolution, or depthwise convolution, is widely used in efficient and lightweight networks design for mobile devices \cite{howard2017mobilenets,zhang2018shufflenet}. Different from the traditional convolution that each kernel applies to all channels of input, channel-wise convolution first divides input channels into $g$ groups and then performs convolution independently over each group. Therefore the computation cost can be reduced by a factor of $g$ compare with the original convolution.
	%-------------------------------------------------------------------------
	
	\section{Approach}
	
	%-------------------------------------------------------------------------
	\begin{figure}[tbp]
		\begin{center}
			\includegraphics[width=\columnwidth]{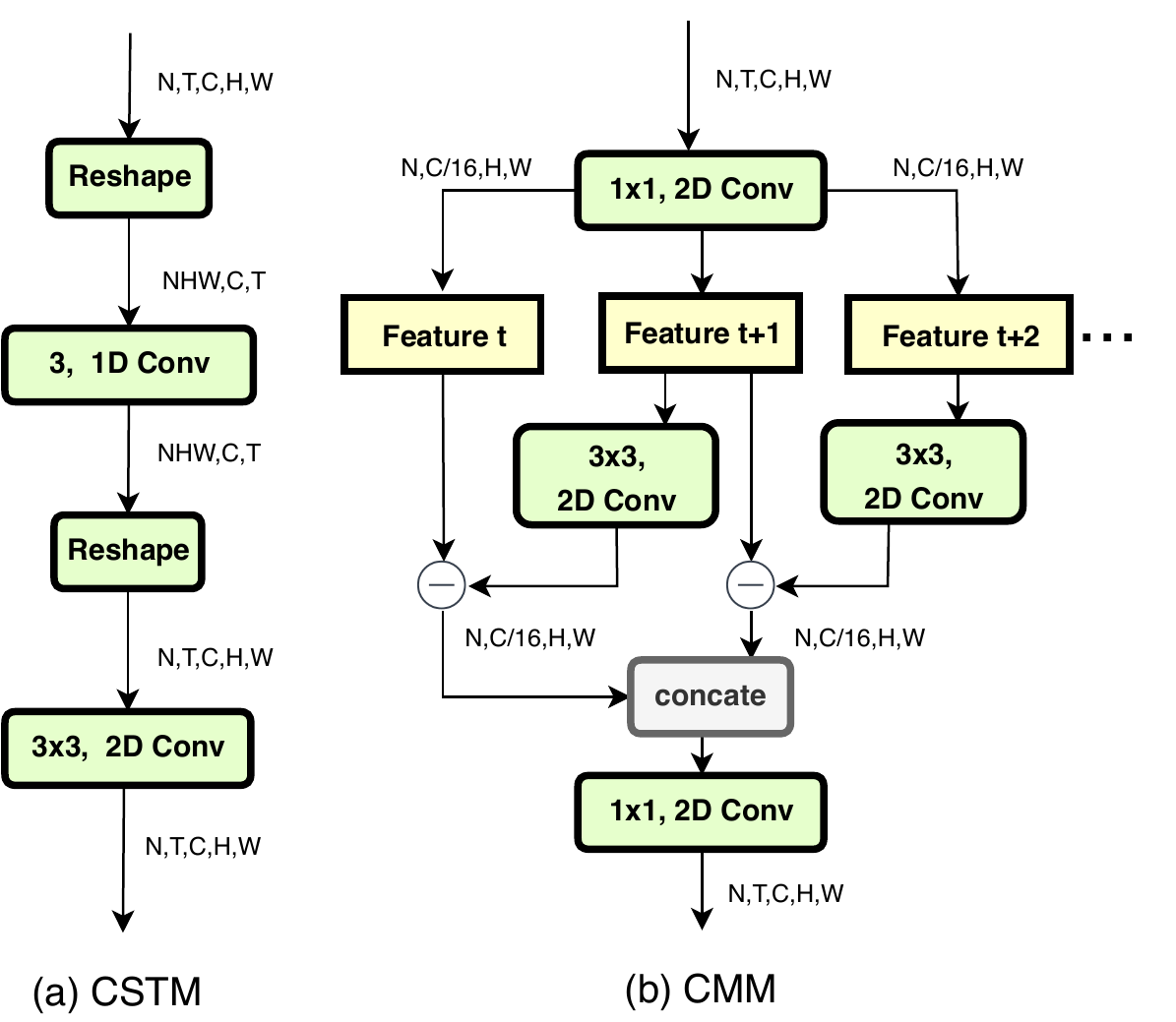}
		\end{center}
		\caption{Architecture of Channel-wise SpatioTemporal Module and Channel-wise Motion Module. The feature maps are shown as the shape of their tensors. "$\ominus$" denotes element-wise subtraction.}
		\label{block}
	\end{figure}
	%-------------------------------------------------------------------------
	%-------------------------------------------------------------------------
	\begin{figure*}[tbp]
		\begin{center}
			\includegraphics[width=0.85\linewidth]{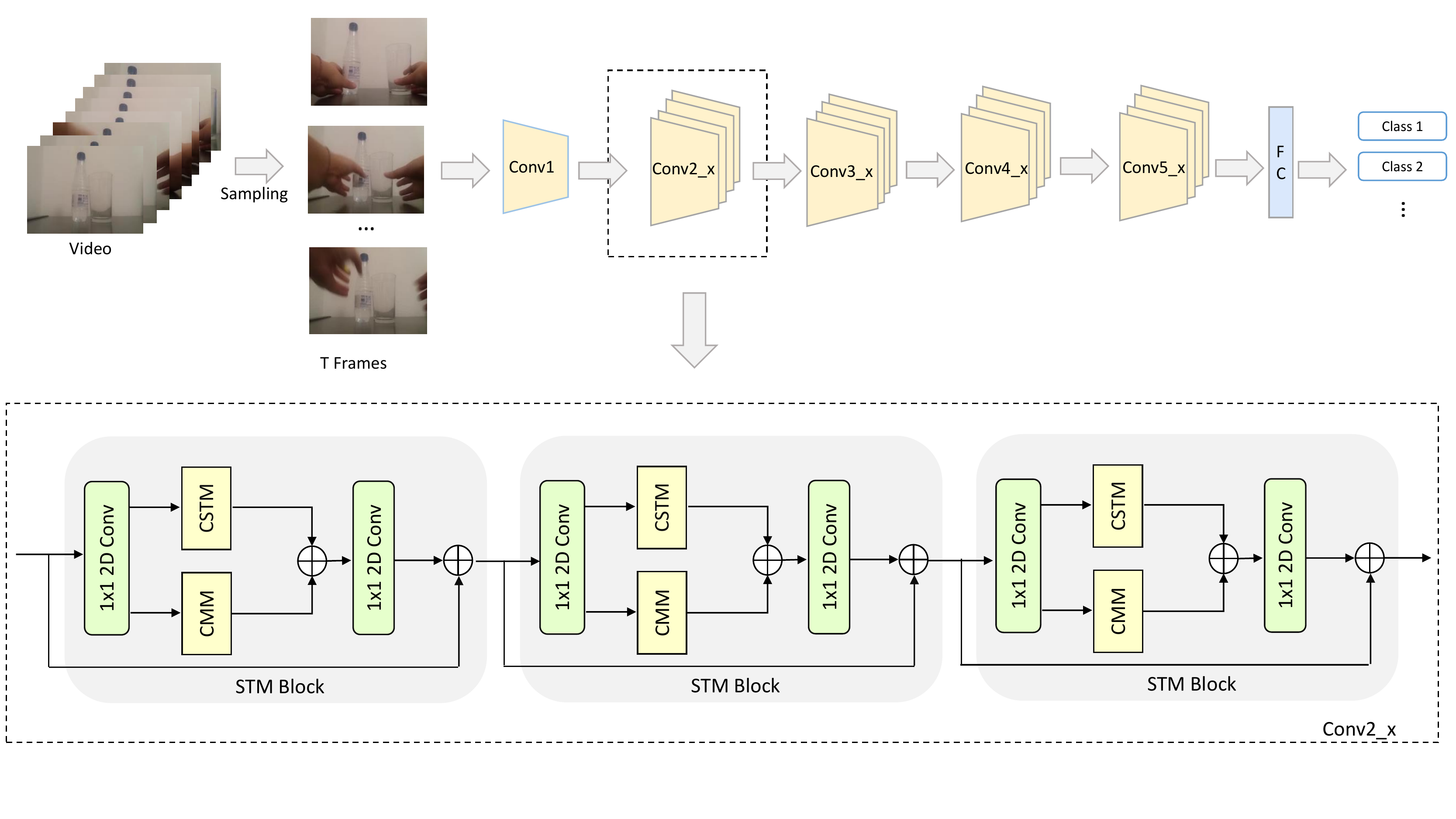}
			\vspace{-0.5cm}
		\end{center}
		\caption{The overall architecture of STM network. The input video is first split into $N$ segments equally and then one frame from each segment is sampled. We adopt 2D ResNet-50 as backbone and replace all residual blocks with STM blocks. No temporal dimension reduction performed apart from the last score fusion stage.}
		\label{overall}
			\vspace{-0.3cm}
	\end{figure*}

	%-------------------------------------------------------------------------
	In this section, we will introduce the technical details of our approach. First, we will describe the proposed CSTM and CMM to show how to perform the channel-wise spatiotemporal fusion and extract the feature-level motion information, respectively. Afterward, we will present the combination of these two modules to assemble them as a building block that can be inserted into existing ResNet architecture to form our STM network.
	
	\subsection{Channel-wise SpatioTemporal Module}
	The CSTM is designed for efficient spatial and temporal modeling. By introducing very limited extra computing cost, CSTM extracts rich spatiotemporal features, which can significantly boost the performance of temporal-related action recognition. As illustrated in Fig. \ref{block}(a), given an input feature map $\mathbf{F} \in \mathbb{R}^{N \times T\times C \times H \times W}$, we first reshape $\mathbf{F}$ as: $\mathbf{F} \rightarrow \mathbf{F}^* \in \mathbb{R}^{NHW \times C \times T}$ and then apply the channel-wise 1D convolution on the \textit{T} dimension to fuse the temporal information. There are mainly two advantages to adopt the channel-wise convolution rather than the ordinary convolution. Firstly, for the feature map $\mathbf{F}^*$, the semantic information of different channels is typically different. We claim that the combination of temporal information for different channels should be different. Thus the channel-wise convolution is adopted to learn independent kernels for each channel. Secondly, compared to the ordinary convolution, the computation cost can be reduced by a factor of $G$ where $G$ is the number of groups. In our settings, $G$ is equal to the number of input channels. Formally, the channel-wise temporal fusion operation can be formulated as:

	\begin{equation}
	\begin{aligned}
	\mathbf{G}_{c,t}=\sum_{i} \mathbf{K}^c_{i}\mathbf{F}^*_{c,t+i}
	\end{aligned}
	\end{equation}
	where $\mathbf{K}^c_{i}$ are temporal combination kernel weights belong to channel $c$ and $i$ is the index of temporal kernel, $\mathbf{F}^*_{c,t+i}$ is the input feature sequence and $\mathbf{G}_{c,t}$ is the updated version of the channel-wise temporal fusion features. Here the temporal kernel size is set to 3 thus $i \in [-1,1]$. Next we will reshape the $\mathbf{G}$ to the original input shape (i.e. $[N,T,C,H,W]$) and model local-spatial information via 2D convolution whose kernel size is 3x3.
	
	We visualize the output feature maps of CSTM to help understand this module in Fig. \ref{feature map}. Compare the features in the second row to the third row, we can find that the CSTM has learned the spatiotemporal features which pay more attention in the main part of the actions such as the hands in the first column while the background features are weak.
	
	\subsection{Channel-wise Motion Module}
	As discovered in \cite{varol2018long,carreira2017quo}, apart from the spatiotemporal features directly learned by 3D CNN from the RGB stream, the performance can still be greatly improved by including an optical-flow motion stream. Therefore, apart from the CSTM, we propose a lightweight Channel-wise Motion Module (CMM) to extract feature-level motion patterns between adjacent frames. Note that our aim is to find the motion representation that can help to recognize actions in an efficient way rather than accurate motion information (optical flow) between two frames. Therefore, we will only use the RGB frames and not involve any pre-computed optical flow.
	
	Given the input feature maps $\mathbf{F} \in \mathbb{R}^{N \times T\times C \times H \times W}$, we will first leverage a 1x1 convolution layer to reduce the spatial channels by a factor of $r$ to ease the computing cost, which is setting to 16 in our experiments. Then we generate feature-level motion information from every two consecutive feature maps. Taking $\mathbf{F}_t$ and $\mathbf{F}_{t+1}$ for example, we first apply 2D channel-wise convolution to $\mathbf{F}_{t+1}$ and then subtracts from $\mathbf{F}_t$ to obtain the approximate motion representation $\mathbf{H}_{t}$:
	
	\begin{equation}
	\begin{aligned}
	\mathbf{H}_{t}=\sum_{i,j}{\mathbf{K}^c_{i,j} \mathbf{F}_{t+1, c,h+i,w+j}}-\mathbf{F}_{t}
	\end{aligned}
	\end{equation}
	where $c,t,h,w$ denote spatial, temporal channel and two spatial dimensions of the feature map respectively and $\mathbf{K}^c_{i,j} $ denotes the $c$-th motion filter with the subscripts $i,j$ denote the spatial indices of the kernel. Here the kernel size is set to $3 \times 3$ thus $i,j \in [-1,1]$.
	
	As shown in Fig. \ref{block}(b), we perform the proposed CMM to every two adjacent feature maps over the temporal dimension, i.e., $\mathbf{F}_t$ and $\mathbf{F}_{t+1}$, $\mathbf{F}_{t+1}$ and $\mathbf{F}_{t+2}$, etc. Therefore, the CMM will produce $T-1$ motion representations. To keep the temporal size compatible with the input feature maps, we simply use zero to represent the motion information of the last time step and then concatenate them together over the temporal dimension. In the end, another 1x1 2D convolution layer is applied to restore the number of channels to $C$.

	We find that the proposed CMM can boost the performance of the whole model even though the design is quite simple, which proves that the motion features obtained with CMM are complementary to the spatiotemporal features from CSTM. We visualize the motion features learned by CMM in Fig. \ref{feature map}. From which we can see that compared to the output of CSTM, CMM  is able to capture the motion features with the distinct edges just like optical flows.
	
	\subsection{STM Network}
	In order to keep the framework effective yet lightweight, we combine the proposed CSTM and CMM together to build an STM block that can encode spatiotemporal and motion features together and can be easily inserted into the existing ResNet architectures. The overall design of the STM block is illustrated in the bottom half of Fig. \ref{overall}. In this STM block, the first 1x1 2D convolution layer is responsible for reducing the channel dimensions. The compressed feature maps are then passed through the CSTM and CMM to extract spatiotemporal and motion features respectively. Typically, there are two kinds of ways to aggregate different type of information: summation and concatenation. We experimentally found that summation works better than concatenation to fuse these two modules. Therefore, an element-wise sum operation is applied after the CSTM and CMM to aggregate the information. Then another 1x1 2D convolution layer is applied to restore the channel dimensions. Similar to the ordinary residual block, we also add a parameter-free identity shortcut from the input to the output.
	
	Because the proposed STM block is compatible with the ordinary residual block, we can simply insert it into any existing ResNet architectures to form our STM network with very limited extra computation cost. We illustrate the overall architecture of STM network in the top half of Figure \ref{overall}. The STM network is a 2D convolutional network which avoids any 3D convolution and pre-computing optical flow. Unless specified, we choose the 2D ResNet-50 \cite{he2016deep} as our backbone for its tradeoff between the accuracy and speed. We replace all residual blocks with the proposed STM blocks.
	
	%-------------------------------------------------------------------------
	\section{Experiments}
	\label{exp}
	In this section, we first introduce the datasets and the implementation details of our proposed approach. Then we perform extensive experiments to demonstrate that the proposed STM outperforms all the state-of-the-art methods on both temporal-related datasets (i.e., Something-Something v1 \& v2 and Jester) and scene-related datasets (i.e., Kinetics-400, UCF-101, and HMDB-51). The baseline method in our experiments is Temporal Segment Networks (TSN) \cite{wang2016temporal} where we replace the backbone to ResNet-50 for fair comparisons. We also conduct abundant ablation studies with Something-Something v1 to analyze the effectiveness of our method. Finally, we give runtime analyses to show the efficiency of STM compare with state-of-the-art methods.
	
	\subsection{Datasets}
\begin{figure}[tbp]
		\begin{center}
			\includegraphics[width=\columnwidth]{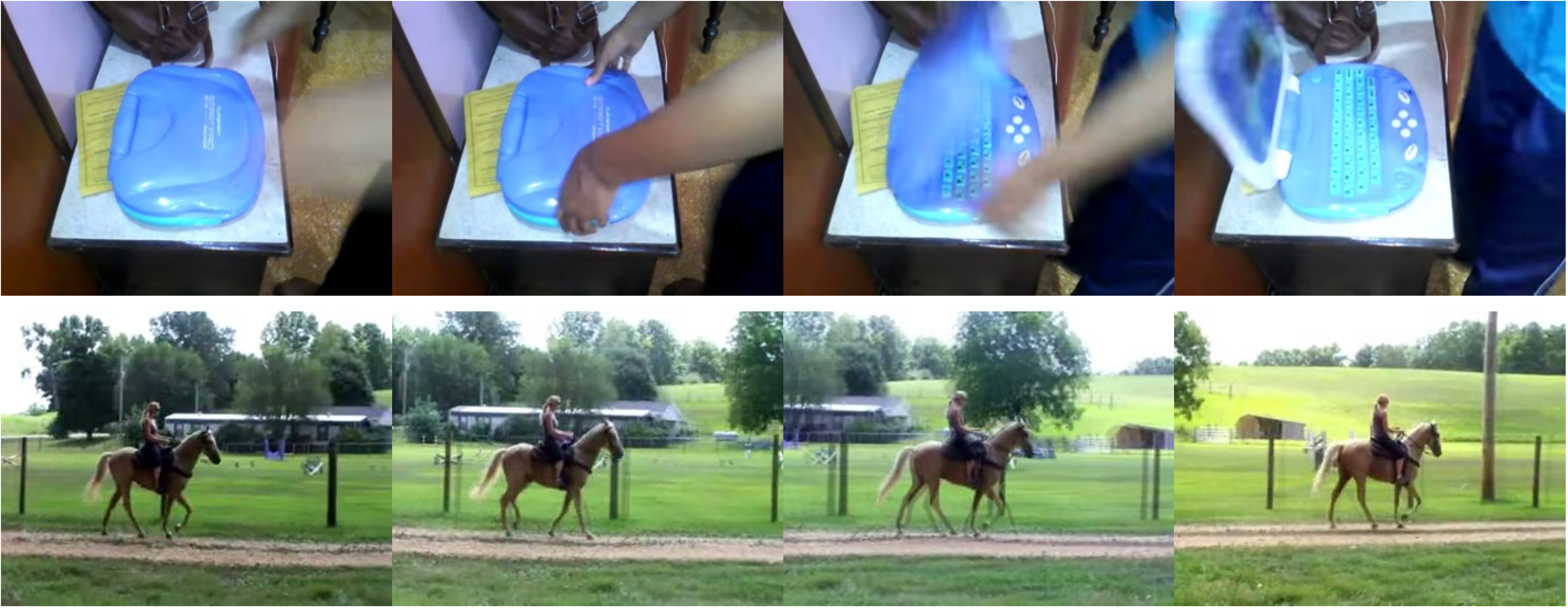}
		\end{center}
		\caption{Difference between temporal-related datasets and scene-related datasets. Top: action for which temporal feature matters. Reversing the order of frames gives the opposite label (opening something vs closing something). Bottom: action for which scene feature matters. Only one frame can predict label (horse riding).}
		\label{dataset}
	\end{figure}

	We evaluate the performance of the proposed STM on several public action recognition datasets. We classify these datasets into two categories: (1) temporal-related datasets, including Something-Something v1 \& v2 \cite{goyal2017something} and Jester \cite{jester}. For these datasets, temporal motion interaction of objects is the key to action understanding. Most of the actions cannot be recognized without considering the temporal relationship; (2)
	scene-related datasets, including Kinetics-400 \cite{carreira2017quo}, UCF-101 \cite{soomro2012ucf101} and HMDB-51 \cite{Kuehne11} where the background information contributes a lot for determining the action label in most of the videos. Temporal relation is not as important as it in the first group of datasets. We also give examples in Figure \ref{dataset} to show the difference between them. Since our method is designed for effective spatiotemporal fusion and motion information extraction, we mainly focus on those temporal-related datasets. Nevertheless, for those scene-related datasets, our method also achieves competitive results.

	%-------------------------------------------------------------------------
	%-------------------------------------------------------------------------
	\begin{table*}[htbp]
		\centering
		\caption{Performance of the STM on the Something-Something v1 and v2 datasets compared with the state-of-the-art methods.}
		\label{some}
		\begin{adjustbox}{max width=\textwidth}
			\begin{threeparttable}
				\begin{tabular}{c|c|c|c|c|c|c|c|c|c|c|c}
					\hline
					\multirow{2}{*}{Method} & \multirow{2}{*}{Backbone}  & \multirow{2}{*}{Flow} & \multirow{2}{*}{Pretrain} & \multirow{2}{*}{Frame} & \multicolumn{3}{c|}{Something-Something v1} & \multicolumn{4}{c}{Something-Something v2}\tabularnewline
					\cline{6-12}
					
					& &  &  &  & top-1 val & top-5 val & top-1 test & top-1 val & top-5 val & top-1 test & top-5 test\tabularnewline
					\hline
					\hline
					S3D-G \cite{xie2018rethinking} & Inception & & ImageNet & 64 & 48.2 & 78.7 & 42.0 & - & - & - & -\tabularnewline
					\hline
					ECO \cite{zolfaghari2018eco} & & & \multirow{4}{*}{Kinetics} & 8 & 39.6 & - & - & - & - & - & -\tabularnewline
					
					ECO \cite{zolfaghari2018eco} & BNInception+ & & & 16 & 41.4 & - & - & - & - & - & -\tabularnewline
					
					ECO$_{EN}Lite$ \cite{zolfaghari2018eco} & 3D ResNet-18 & & & 92 & 46.4 & - & 42.3 & - & - & - & -\tabularnewline
					
					ECO$_{EN}Lite$ Two-Stream \cite{zolfaghari2018eco} &  &  \checkmark & & 92+92 & 49.5 & - & 43.9 & - & - & - & -\tabularnewline
					\hline
					I3D \cite{carreira2017quo} & \multirow{2}{*}{3D ResNet-50}  & & \multirow{2}{*}{Kinetics} & 32 & 41.6 & 72.2 & - & - & - & - & -\tabularnewline
					I3D+GCN \cite{carreira2017quo} & & &   & 32 & 43.4 & 75.1 & - & - & - & - & -\tabularnewline
					\hline
					\hline
					\multirow{2}{*}{TSN \cite{wang2016temporal}} & \multirow{2}{*}{ResNet-50} &\multirow{2}{*}{} & \multirow{2}{*}{Kinetics} & 8 & 19.7 & 46.6 & - & 27.8 & 57.6 & - & -\tabularnewline
					& & & & 16 & 19.9 & 47.3 & - & 30.0 & 60.5 & - & -\tabularnewline
					\hline
					TRN Multiscale \cite{zhou2018temporal} & \multirow{2}{*}{BNInception} &  & \multirow{2}{*}{ImageNet} & 8 & 34.4 & - & 33.6 & 48.8 & 77.64 & 50.9 & 79.3\tabularnewline
					
					TRN Two-Stream \cite{zhou2018temporal} &  & \checkmark & & 8+8 & 42.0 & - & 40.7 & 55.5 & 83.1 & 56.2 & 83.2\tabularnewline
					\hline
					MFNet-C101 \cite{lee2018motion} & ResNet-101 & & Scratch & 10 & 43.9 & 73.1 & 37.5 & - & - & - & -\tabularnewline
					\hline
					TSM \cite{lin2018temporal} & \multirow{2}{*}{ResNet-50} & & \multirow{2}{*}{Kinetics} & 16 & 44.8 & 74.5 & - & 58.7 & 84.8 & 59.9 & 85.9\tabularnewline
					TSM Two-Stream \cite{lin2018temporal} &   & \checkmark&  & 16+8 & 49.6 & 79.0 & \textbf{46.1} & 63.5 & 88.6 & \textbf{63.7} & 89.5\tabularnewline
					\hline
					\multirow{2}{*}{\textbf{STM}} & \multirow{2}{*}{ResNet-50} & & \multirow{2}{*}{ImageNet} & 8 & 49.2 & 79.3 & - & 62.3 & 88.8 & 61.3 & 88.4\tabularnewline
					&  & &   & 16 & \textbf{50.7} & \textbf{80.4} & 43.1 & \textbf{64.2} & \textbf{89.8} & 63.5 & \textbf{89.6}\tabularnewline
					\hline
				\end{tabular}%
			\end{threeparttable}
		\end{adjustbox}
	\end{table*}

	\begin{table}[htbp]
		\centering
		\caption{Performance of the STM on the Jester compared with the state-of-the-art methods.}
		\label{jester}
		\begin{adjustbox}{max width=\columnwidth}
			\begin{threeparttable}
				\begin{tabular}{c|c|c|c|c}
					\hline
					Method & Backbone & Frame & Top-1 & Top-5\tabularnewline
					\hline
					\hline
					\multirow{2}{*}{TSN \cite{wang2016temporal}} & \multirow{2}{*}{ResNet-50} & 8 & 81.0 & 99.0\tabularnewline
					
					&   & 16 & 82.3 & 99.2\tabularnewline
					\hline
					TRN-Multiscale \cite{zhou2018temporal} & BNInception & 8 & 95.3 & -\tabularnewline
					\hline
					MFNet-C50 \cite{lee2018motion} & ResNet-50 & 7 & 96.1 & 99.7\tabularnewline
					\hline
					\multirow{2}{*}{TSM \cite{lin2018temporal}} & \multirow{2}{*}{ResNet-50} & 8 & 94.4 & 99.7\tabularnewline
					&  & 16 & 95.3 & 99.8\tabularnewline
					\hline
					\multirow{2}{*}{\textbf{STM}} & \multirow{2}{*}{ResNet-50} & 8 & 96.6 & 99.9\tabularnewline
					&  & 16 & \textbf{96.7} & \textbf{99.9}\tabularnewline
					\hline
				\end{tabular}%
			\end{threeparttable}
		\end{adjustbox}
	\end{table}
	%若篇幅不够，这里可以再补充数据集介绍和用的指标等
	
	\subsection{Implementation Details}
	\textbf{Training.} We train our STM network with the same strategy as mentioned in TSN \cite{wang2016temporal}. Given an input video, we first divide it into $T$ segments of equal durations in order to conduct long-range temporal structure modeling. Then, we randomly sample one frame from each segment to obtain the input sequence with $T$ frames. The size of the short side of these frames is fixed to 256. Meanwhile,  corner cropping and scale-jittering are applied for data argumentation. Finally, we resize the cropped regions to 224$\times$224 for network training. Therefore, the input size of the network is $N\times T \times 3 \times 224 \times 224$, where $N$ is the batch size and $T$ is the number of the sampled frames per video. In our experiments, $T$ is set to 8 or 16.

	We train our model with 8 GTX 1080TI GPUs and each GPU processes a mini-batch of 8 video clips (when $T=8$) or 4 video clips (when $T=16$). For Kinetics, Something-Something v1 \& v2 and Jester, we start with a learning rate of 0.01 and reduce it by a factor of 10 at 30,40,45 epochs and stop at 50 epochs. For these large-scale datasets, we only use the ImageNet pre-trained model as initialization. For the temporal channel-wise 1D convolution in CSTM, first quarter of channels are initialized to [1,0,0], last quarter of channels are initialized to [0,0,1] and other half are [0,1,0]. All parameters in CMM are randomly initialized. For UCF-101 and HMDB-51, we use Kinetics pre-trained model as initialization and start training with a learning rate of 0.001 for 25 epochs. The learning rate is decayed by a factor 10 every 15 epochs. We use mini-batch SGD as optimizer with a momentum of 0.9 and a weight decay of 5e-4. Different from \cite{wang2016temporal}, we enable all the BatchNorm layers \cite{ioffe2015batch} during training.

	\textbf{Inference.} Following \cite{wang2018non,feichtenhofer2018slowfast}, we first scale the shorter spatial side to 256 pixels and take three crops of $256 \times 256$ to cover the spatial dimensions and then resize them to $224 \times 224$. For the temporal domain, we randomly sample 10 times from the full-length video and compute the softmax scores individually. The final prediction is the averaged softmax scores of all clips.

	\subsection{Results on Temporal-Related Datasets}
	In this section, we compare our approach with the state-of-the-art methods on temporal-related datasets including Something-Something v1 \& v2 and Jester. Something-Something v1 is a large collection of densely-labeled video clips which shows basic human interactions with daily objects. This dataset contains 174 classes with 108,499 videos. Something-Something v2 is an updated version of v1 with more videos (220,847 in total) and greatly reduced label noise. Jester is a crowd-acted video dataset for generic human hand gestures recognition, which contains 27 classes with 148,092 videos.
	\begin{table}[tbp]
	\scriptsize
	\centering
	\caption{Performance of the STM on the Kinetics-400 dataset compared with the state-of-the-art methods.}
	\label{kinetics}
	\begin{adjustbox}{max width=\columnwidth}
		\begin{threeparttable}
\begin{tabular}{c|c|c|c|c}
\hline
Method  & Backbone  & Flow  & Top-1  & Top-5\tabularnewline
\hline
\hline
STC \cite{diba2018spatio}  & ResNext101  &  & 68.7  & 88.5\tabularnewline
\hline
ARTNet \cite{wang2018appearance}  & ResNet-18  &  & 69.2  & 88.3\tabularnewline
\hline
\multirow{2}{*}{ECO \cite{zolfaghari2018eco}}  & BNInception  &  & \multirow{2}{*}{70.7 }  & \multirow{2}{*}{89.4}\tabularnewline
 & +3D ResNet-18  &  &  & \tabularnewline
\hline
S3D \cite{xie2018rethinking}  & Inception  &  & 72.2  & 90.6\tabularnewline
\hline
I3D RGB \cite{carreira2017quo}  & \multirow{2}{*}{3D Inception-v1}  &  & 71.1  & 89.3\tabularnewline
I3D Two-Stream \cite{carreira2017quo}  &  & \checkmark  & \textbf{74.2}  & 91.3\tabularnewline
\hline
StNet \cite{he2018stnet}  & ResNet-101  &  & 71.4  & -\tabularnewline
\hline
Disentangling \cite{zhao2018recognize}  & BNInception  &  & 71.5  & 89.9\tabularnewline
\hline
R(2+1)D RGB \cite{tran2018closer}  & \multirow{2}{*}{ResNet-34}  &  & 72.0  & 90.0\tabularnewline
R(2+1)D Two-Stream \cite{tran2018closer}  &  & \checkmark  & 73.9  & 90.9\tabularnewline
\hline
TSM \cite{lin2018temporal} & ResNet-50 &  & 72.5 & 90.7\tabularnewline
\hline
TSN RGB \cite{wang2016temporal}  & \multirow{2}{*}{BNInception}  &  & 69.1  & 88.7\tabularnewline
TSN Two-Stream \cite{wang2016temporal}  &  & \checkmark  & 73.9  & 91.1\tabularnewline
\hline
\textbf{STM}  & ResNet-50  &  & 73.7  & \textbf{91.6}\tabularnewline
\hline
\end{tabular}
		\end{threeparttable}
	\end{adjustbox}
\vspace{-0.3cm}
	\end{table}
	
		\begin{table*}[tbp]
		\centering
		\caption{Performance of the STM on UCF-101 and HMDB-51 compared with the state-of-the-art methods.}
		\label{UCFHMDB}
		\begin{adjustbox}{max width=0.98\textwidth}
			\begin{threeparttable}
\begin{tabular}{c|c|c|c|c|c}
\hline
Method  & Backbone  & Flow  & Pre-train Data  & UCF-101  & HMDB-51\tabularnewline
\hline
\hline
C3D \cite{tran2015learning}  & 3D VGG-11 &  & Sports-1M  & 82.3  & 51.6\tabularnewline
\hline
STC \cite{diba2018spatio}  & ResNet101  &  & Kinetics  & 93.7  & 66.8\tabularnewline
\hline
ARTNet with TSN \cite{wang2018appearance}  & 3D ResNet-18  &  & Kinetics  & 94.3  & 70.9\tabularnewline
\hline
ECO \cite{zolfaghari2018eco}  & BNInception+3D ResNet-18  &  & Kinetics  & 94.8  & 72.4\tabularnewline
\hline
I3D RGB \cite{carreira2017quo}  & \multirow{2}{*}{3D Inception-v1}  &  & \multirow{2}{*}{ImageNet+Kinetics}  & 95.1  & 74.3\tabularnewline
I3D two-stream \cite{carreira2017quo}  &  & \checkmark  &  & 98.0  & 80.7\tabularnewline
\hline
\hline
TSN \cite{wang2016temporal}  & ResNet-50  &  & ImageNet  & 86.2  & 54.7\tabularnewline
\hline
TSN RGB \cite{wang2016temporal}  & \multirow{2}{*}{BNInception}  &  & \multirow{2}{*}{ImageNet+Kinetics}  & 91.1  & -\tabularnewline
TSN two-Stream \cite{wang2016temporal}  &  & \checkmark  &  & 97.0  & -\tabularnewline
\hline
TSM \cite{lin2018temporal} & ResNet-50 &  & ImageNet+Kinetics & 94.5 & 70.7\tabularnewline
\hline
StNet \cite{he2018stnet}  & ResNet50  &  & ImageNet+Kinetics & 93.5  & -\tabularnewline
\hline
Disentangling \cite{zhao2018recognize}  & BNInception  &  & ImageNet+Kinetics & 95.9  & -\tabularnewline
\hline
\textbf{STM}  & ResNet-50  &  & ImageNet+Kinetics & 96.2  & 72.2\tabularnewline
\hline
\end{tabular}
			\end{threeparttable}
		\end{adjustbox}
\vspace{-0.3cm}
	\end{table*}
	
	Table \ref{some} lists the results of our method compared with the state-of-the-art on Something-Something v1 and v2. The results of the baseline method TSN are relatively low compared with other methods, which demonstrates the importance of temporal modeling for these temporal-related datasets. Compared with the baseline method, our STM network gains 29.5\% and 30.8\% top-1 accuracy improvement with 8 and 16 frames inputs respectively on Something-Something v1. On Something-Something v2, STM also gains 34.5\% and 34.2\% improvement compared to TSN. The rest part of Table \ref{some} shows the other state-of-the-art methods. These methods can be classified into two types as shown in the two parts of Table \ref{some}. The upper part presents the 3D CNN based methods, including S3D-G \cite{xie2018rethinking}, ECO \cite{zolfaghari2018eco} and I3D+GCN models \cite{wang2018videos}. The lower part is 2D CNN based methods, including TRN \cite{zhou2018temporal}, MFNet \cite{lee2018motion} and TSM \cite{lin2018temporal}. It is clear that even STM with 8 RGB frames as input achieves the state-of-the-art performance compared with other methods, which take more frames and optical flow as input or 3D CNN as the backbone. With 16 frames as input, STM achieves the best performance in the validation sets of both Something-Something v1 and v2, and just a little lower in the top1 accuracy in the test sets, which adopts only 16 RGB frames as input.

	Table \ref{jester} shows the results on the Jester dataset. Our STM also gains a large improvement compared to the TSN baseline method, and outperforms all the state-of-the-art methods.
	%------------------------------------------------------------------------

	\subsection{Results on Scene-Related Datasets}
	We evaluate our STM on three scene-related datasets: Kinetics-400, UCF-101, and HMDB-51 in this section. Kinetics-400 is a large-scale human action video dataset with 400 classes. It contains 236,763 clips for training and 19,095 clips for validation. UCF-101 is a relatively small dataset which contains 101 categories and 13,320 clips in total. HMDB-51 is also a small video dataset with 51 classes and 6766 labeled video clips. For UCF-101 and HMDB-51, we followed \cite{wang2016temporal} to adopt the three training/testing splits for evaluation.

	Table \ref{kinetics} summaries the results of STM and other competing methods on the Kinetics-400 dataset. We train STM with 16 frames as input, and the same for evaluation. From the evaluation results, we can draw the following conclusions: (1) Different from the previous temporal-related datasets, most actions of Kinetics can be recognized by scene and objects even with one still frame of videos, therefore the baseline method without any temporal modeling can achieve acceptable accuracy; (2) Though our method is mainly focused on temporal-related actions recognition, STM still achieves very competitive results compare with the state-of-the-art methods. Top-1 accuracy of our method is only 0.5\% lower than the two-stream I3D, which involves both 3D convolution and pre-computation optical flow. However, STM outperforms major recently proposed 3D CNN based methods (the upper part of the Table \ref{kinetics}) as well as 2D CNN based methods (the lower part of the Table \ref{kinetics}) and achieve the best top-5 accuracy compared with all the other method.
	%-------------------------------------------------------------------------	

	We also conduct experiments on the UCF-101 and HMDB-51 to study the generalization ability of learned spatiotemporal and motion representations. We evaluate our method over three splits and report the averaged results in Table \ref{UCFHMDB}. First, compared with the ImageNet pre-trained model, Kinetics pre-train can significantly improve the performance on small datasets. Then, compare with the state-of-the-art methods, only two methods,  I3D two-stream and TSN two-Stream, performs a little better than ours while both of them utilize optical flow as their extra inputs. However, STM with 16 frames as inputs even outperforms I3D with RGB stream on UCF101, which also uses Kinetics as pre-train data but the 3D CNN leads to much higher computation cost than ours.
	
\subsection{Ablation Studies}
In this section, we comprehensively evaluate our proposed STM on Something-Something v1 dataset. All the ablation experiments in this section use 8 RGB frames as inputs.

\begin{table*}[t]
\begin{minipage}[t]{0.5\columnwidth}	
\vspace{0pt}
	\begin{adjustbox}{max width=\columnwidth,valign=t}
		\begin{threeparttable}
			\begin{tabular}{c|c|c}
				\hline
				Model & Top-1 & Top-5\tabularnewline
				\hline
				\hline
				TSN & 19.7 & 46.6\tabularnewline
				\hline
				CSTM & 47.7 & 77.9\tabularnewline
				
				CMM & 44.1 & 74.8\tabularnewline
				
				STM & 49.2 & 79.3\tabularnewline
				\hline
			\end{tabular}
		\end{threeparttable}
	\end{adjustbox}
\vspace{1pt}
	\caption{\textbf{Impact of two modules:} Comparison between CSTM, CMM and STM.}
\label{ablation_tables1}
\end{minipage}
\begin{minipage}[t]{0.5\columnwidth}
\vspace{0pt}
	\begin{adjustbox}{max width=\columnwidth,valign=t}
		\begin{threeparttable}
			\begin{tabular}{c|c|c}
				\hline
				Aggregation & Top-1 & Top-5\tabularnewline
				\hline
				\hline
				TSN & 19.7 & 46.6\tabularnewline
				\hline
				Summation & 49.2 & 79.3\tabularnewline
				Concatenation & 41.8 & 73.2\tabularnewline
				\hline
			\end{tabular}
		\end{threeparttable}
	\end{adjustbox}
\vspace{17pt}
\caption{\textbf{Fusion of two modules:} Summation fusion is better.}
\label{ablation_tables2}
\end{minipage}
\begin{minipage}[t]{0.5\columnwidth}
\vspace{0pt}
	\begin{adjustbox}{max width=\columnwidth,valign=t}
		\begin{threeparttable}
			\begin{tabular}{c|c|c|c}
				\hline
				Stage & STM Blocks & Top-1 & Top-5\tabularnewline
				\hline
				\hline
				2 & 1 & 38.7 & 70.1\tabularnewline
				3 & 1 & 40.6 & 71.6\tabularnewline
				4 & 1 & 41.5 & 72.6\tabularnewline
				5 & 1 & 41.5 & 71.8\tabularnewline
				\hline
				2-5 & 4 & 47.9 & 78.1\tabularnewline
				\hline
				2-5 & 16 & 49.2 & 79.3\tabularnewline
				\hline
			\end{tabular}
		\end{threeparttable}
	\end{adjustbox}
    \vspace{0pt}
	\caption{\textbf{Location and number of STM block:} Deeper location and more blocks yeild better performance.}
\label{ablation_tables3}
\end{minipage}
\begin{minipage}[t]{0.5\columnwidth}
\vspace{0pt}
	\begin{adjustbox}{max width=\linewidth,valign=t}
		\begin{threeparttable}[t]
            \begin{tabular}[t]{c|c|c}
            \hline
            Type & Channel-wise & Ordinary\tabularnewline
            \hline
            \hline
            Top-1 Acc. & 47.7 & 46.9\tabularnewline
            \hline
            Param. & 23.88M & 27.64M\tabularnewline
            \hline
            FLOPs & 32.93G & 40.59G\tabularnewline
            \hline
            \end{tabular}
		\end{threeparttable}
	\end{adjustbox}
    \vspace{27pt}

	\caption{\textbf{Type of temporal convolution in CSTM:} Channel-wise temporal convolution yields better performance.}
    \label{ablation_tables4}
\end{minipage}
\end{table*}

	\noindent \textbf{Impact of two modules.} Our proposed two modules can be inserted into a standard ResNet architecture independently. To validate the contributions of each component in the STM block (i.e., CSTM and CMM), we compare the results of the individual module and the combination of both modules in Table \ref{ablation_tables1}. We can see that each component contributes to the proposed STM block. CSTM learns channel-wise temporal fusion and brings about 28\% top-1 accuracy improvement compared to the baseline method TSN while CMM learns feature-level motion information and brings 24.4\% top-1 accuracy improvement. When combining CSTM and CMM together, we can learn richer spatiotemporal and motion features and achieve the best top-1 accuracy, especially, the gain over the baseline is 29.5\%.

\setlength{\parskip}{0pt}
	\noindent \textbf{Fusion of two modules.} There are two ways to combine CSTM and CMM: element-wise summation and concatenation. The element-wise summation is parameter-free and easy to implement. For concatenation fusion, we first concatenate outputs of CSTM and CMM over the channel dimension, and the dimension of concatenate features is $2C$. Then a 1x1 convolution is applied to reduce the channels to $C$. We conduct the experiments to study the two fusion ways as shown in Table \ref{ablation_tables2}, though summation aggregation is simple, it still outperforms concatenation by 7.4\% at top-1 accuracy and 6.1\% at top-5 accuracy.

\setlength{\parskip}{0pt}
	\noindent \textbf{Location and number of STM block.} ResNet-50 architecture can be divided into 6 stages. We refer the conv2\_x to conv5\_x as stage 2 to stage 5. The first four rows of Table \ref{ablation_tables3} compare the performance of replacing only the first residual block with STM on different stages in ResNet-50, from stage 2 to stage 5, respectively. We conclude from the results that replacing only one residual block already yield significant performance improvement compared to the baseline TSN, which demonstrates the effectiveness of the proposed STM block. One may notice that replacing the STM block at latter stage (e.g., stage 5) yield better accuracy than early stage (e.g., stage 2). One possible reason is that temporal modeling is beneficial more with larger receptive fields which can capture holistic features. We then replace one block for each stage (i.e., replacing four blocks in all) and leads to better results. When replacing all original residual blocks with STM blocks (i.e., 16 blocks in all), our model achieves the best performance.

    \noindent \textbf{Type of temporal convolution in CSTM.} We choose channel-wise temporal convolution in CSTM to learn temporal combination individually for each channel. We also make comparison with ordinary temporal convolution in CSTM module and the result is shown in Table \ref{ablation_tables4}. With channel-wise convolution, we can achieve better performance with few parameters and FLOPs.

	%-------------------------------------------------------------------------

	\subsection{Runtime Analysis}
	Our STM achieves the new state-of-the-art results on several benchmark datasets compared with other methods. More importantly, it is a unified 2D CNN framework without any time-consuming 3D convolution and optical flow calculations. Table \ref{runtime} shows the accuracy and model complexity of STM and several state-of-the-art methods on Something-Something v1 dataset. All evaluations are running on one GTX 1080TI GPU. For a fair comparison, we evaluate our method by evenly sampling 8 or 16 frames from a video and then apply the center crop. To evaluate speed, we use a batch size of 16 and ignore the time of data loading. Compared to I3D and ECO, STM achieves approximately 10x and 2x less FLOPs (33.3G vs 306G, 64G) while 5.9\% and 6.1\% higher accuracy.
%One may notice that our method is somewhat slower than TSM with almost the same FLOPs. The reason is that the widely used channel-wise convolution in our model has not been well optimized in the current version of CUDA and CuDNN. Therefore, our model can still gain a higher speedup with a well-optimized channel-wise convolution kernel. Even so,
Compared to $\rm TSM_{16F}$, our $\rm STM_{8F}$ gains 2.7\% higher accuracy with 1.5x faster speed and half FLOPs.

	\begin{table}[tbp]
		\centering
		\caption{Accuracy and model complexity of STM and other state-of-the-art methods on Something-Something V1 dataset. Single crop STM beats all competing methods with 62 videos per second with 8 frames as input. Measured on a single NVIDIA GTX 1080TI GPU.}
		\label{runtime}
		\begin{adjustbox}{max width=\textwidth}
			\begin{threeparttable}
				\begin{tabular}{c|c|c|c|c|c}
					\hline
					Model & Frame & FLOPs & Param. & Speed & Acc.\tabularnewline
					\hline
					\hline
					I3D \cite{carreira2017quo} & 64 & 306G & 28.0M & 6.4 V/s & 41.6\tabularnewline
					\hline
					ECO \cite{zolfaghari2018eco} & 16 & 64G & 47.5M & 46.3 V/s & 41.4\tabularnewline
					\hline
					\multirow{2}{*}{TSM \cite{lin2018temporal}} & 8 & 32.9G & \multirow{2}{*}{23.9M} & 80.4 V/s & 43.8\tabularnewline
					\cline{2-3} \cline{3-3} \cline{5-6} \cline{6-6}
					& 16 & 65.8G &  & 40.6 V/s & 44.8\tabularnewline
					\hline
					\multirow{2}{*}{\textbf{STM}} & 8 & 33.3G & \multirow{2}{*}{24.0M} & 62.0 V/s & 47.5\tabularnewline
					\cline{2-3} \cline{3-3} \cline{5-6} \cline{6-6}
					& 16 & 66.5G &  & 32.0 V/s & \textbf{49.8}\tabularnewline
					\hline
				\end{tabular}
			\end{threeparttable}
		\end{adjustbox}
	\end{table}
	
	%-------------------------------------------------------------------------
\section{Conclusion}
	In this paper, we presented a simple yet effective network for action recognition by encoding spatiotemporal and motion features together in a unified 2D CNN network. We replace the original residual blocks with STM blocks in ResNet architecture to build the STM network. An STM block contains a CSTM to model channel-wise spatiotemporal feature and a CMM to model channel-wise motion representation together. Without any 3D convolution and pre-calculation optical flow, our STM receives state-of-the-art results on both temporal-related datasets and scene-related datasets with only 1.2\% more FLOPs compared to TSN baseline.
	
{\small
	\bibliographystyle{ieee_fullname}
	\bibliography{egbib}
}

\end{document}